\documentclass[runningheads]{llncs}

\usepackage{graphicx}
\usepackage{url}
\usepackage{multirow}

\begin{document}

\title{\textit{ProtagonistTagger} -- a Tool for Entity Linkage of Persons in Texts from Various Languages and Domains
\thanks{Partially funded by NCBiR, POLNOR, NOR/POLNOR/REFSA/0059/2019.}
}
\author{Weronika Łajewska\inst{1} \and Anna Wróblewska\inst{1}\orcidID{0000-0002-3407-7570} }

\authorrunning{W. Łajewska et al.}

\institute{Warsaw University of Technology, Poland \\
\email{weronikalajewska@gmail.com,anna.wroblewska1@pw.edu.pl}}

\maketitle

\begin{abstract}
Named entities recognition (NER) and disambiguation (NED) can add semantic context to the recognized named entities in texts. Named entity linkage in texts, regardless of a domain, provides links between the entities mentioned in unstructured texts and individual instances of real-world objects. In this poster, we present a tool -- \textit{protagonistTagger} -- for \textit{person} NER and NED in texts. The tool was tested on texts extracted from classic English novels and Polish Internet news. The tool's performance (both precision and recall) fluctuates between 78\% and even 88\%.

\keywords{Named Entity Recognition \and Named Entity Disambiguation \and Entity Linkage \and Semantic Similarity \and Approximate String Matching.}
\end{abstract}

\section{Introduction}
Extracting, integrating and matching instances of named entities referring to the same real-world objects remains a big challenge in the field of Natural Language Processing and the Semantic Web. Nevertheless, this task is necessary to achieve the understanding and integrity of various resources. One of the most basic and standard categories of named entities appearing in almost every type of text is referring to people. Named entity recognition (NER) models can provide us with annotations that are undifferentiated as they are all tagged with general label \emph{person}. In order to be able to analyze text on deeper levels or find shared factors between various texts, we need contradistinction between the recognized \textit{person} named entities. The most desired way is to have a unique identifier for each person appearing in the considered dataset. This identifier can be the full name of the recognized person. Ideally, each person's mention in the dataset should be linked with a tag containing the full name of the corresponding person. 

Up to our best knowledge, there are no available tools that combine the recognition and identification of persons in various domains. There are available NER models, particularly for news and common language texts. Search engines can search through particular names and match distinct names each time we query them. However, it is complicated to recognize and match mentions of specific people with named entities in an arbitrary text using a single tool. 

\begin{figure}
    \centering
    \includegraphics[width=\textwidth]{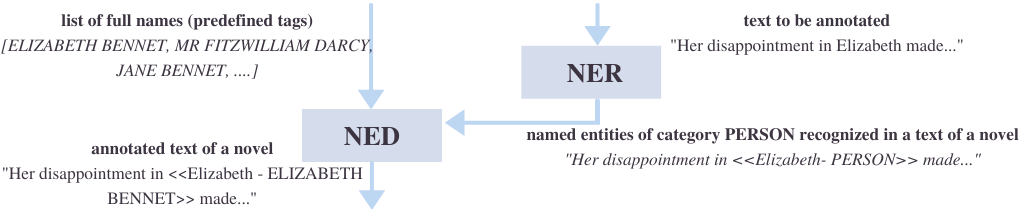}
    \caption{Simplified process employed in our tool -- \emph{protagonistTagger} -- with examples.}
    \label{fig:workflow}
\end{figure}

The general workflow of the created tool -- called \emph{protagonistTagger} -- is presented in~Figure~\ref{fig:workflow}. It works automatically with a list of peoples' full names and a text to be annotated given as inputs. The process of \textit{person} entity linkage implemented in \textit{protagonistTagger} is divided into two main phases:
\begin{enumerate}
    \item named entity recognition (NER) of mentions of people in a text.
    \item named entity disambiguation (NED) providing links between the entities recognized in unstructured text and identifiers referring to real-world objects.
\end{enumerate}

\section{\textit{ProtagonistTagger}: Original English Literature Use Case \protect\footnote{The \textit{protagonistTagger} tool, benchmark datasets from literary domain and annotated corpus are available at \protect\url{https://zenodo.org/record/4699418}}}
The idea for creating a tool for named entity linkage dedicated for novels arose from the complexity of texts in the literary domain and the weak performance of standard methods on such complex texts~\cite{detecting_characters}. The novel is a particular type of text in terms of writing style, the links between sentences, the plot's complexity, the number of characters, etc. \par

NER phase of the linkage process in the literary domain uses a pretrained standard NER model fine-tuned with data from literary domain annotated with general tag \textit{person} in a semi-automatic way~\cite{ner_fine-tuning}. It was necessary due to the relatively low performance of standard models trained primarily on web data such as blogs, news, and comments \cite{spacy_eval,spacy_eval_2}. In the NER phase we want to find as many potential \emph{person} entities to be matched in the NED phase as possible (have the highest possible recall). The phase of NED aims at linking each mention of a protagonist in a given text with a proper tag (i.e., a full name of this protagonist), having been given the list of protagonists' proper names predefined for each novel. The matching method is mainly based on \emph{approximate text matching}. The algorithm also incorporates a set of hand-crafted rules (lexico-semantic patterns) for distinguishing between different instances of concept person and it uses several external dictionaries. This way entities' similarities are considered on the semantic and syntactic level. The \textit{protagonisTagger} addresses the problem of diminutives of basic forms of names and disambiguation of people with the same surname (by analyzing personal title preceding surname in a text). \par

The tool was used to prepare a corpus of 13 novels (altogether more than 50,000 sentences) and more than 35,000 annotations of literary characters. The corpus was used for the analysis of sentiment and relationships in novels. The performed analysis was much more manageable and precise, thanks to the available annotations of literary characters.\par

\section{\textit{ProtagonistTagger}: Internet News Use Case \protect\footnote{The adapted tool along with the new datasets: \protect\url{https://zenodo.org/record/5060232}}}
In order to verify the usability of the created tool in a brand new domain, we investigated its performance on a set made from internet news written in Polish~\cite{pachocki2020categorization}. The \textit{protagonistTagger} turned out to be universal, and it was easily adapted for new data. The only required modification is connected with changing the language model from English to Polish. In the NER phase, we skipped the step of fine-tuning the standard NER model. This decision was caused by the relatively high performance of the standard NER model on the news compared with literary texts. Since external resources, such as dictionaries and syntactic rules, are language-dependent, they were disabled in this experiment. The NED phase was primarily based on the \textit{approximate string matching} part of the \textit{matching algorithm}.

\section{Evaluation and Datasets}
Testing sets for the literary domain contain sentences chosen randomly from 13 novels differing in style and genre (\textit{large} -- testing sets from 10 novels, and \textit{small} -- from distinct 3 novels). The testing sets used for \textit{protagonistTagger} contain all together 1,300 sentences (100 sentences from each novel) annotated manually with a general tag \textit{person} for NER testing (see~Table~\ref{tab:metrics_fine_tuned_ner}) and full names of the mentioned people for NED testing, as well as for the testing of the entire tool. 

\begin{table}[!ht]
\caption{Metrics computed for the standard NER models (\textit{en\_core\_web\_sm} or \textit{pl\_core\_news\_sm} available at \protect\url{https://spacy.io/models}) and the fine-tuned NER model for annotations with general label \emph{person}.}
 \scriptsize
 \centering
 \setlength{\arrayrulewidth}{0.1mm}
\setlength{\tabcolsep}{3pt}
\renewcommand{\arraystretch}{1}
\begin{tabular}{ p{3cm} p{2cm} p{1.5cm}  p{1.5cm}  p{1.5cm}  p{1.5cm}}
%\hline
 \multicolumn{2}{c}{\textbf{Testing set/NER model}} & \textbf{Precision} & \textbf{Recall} & \textbf{F-measure} & \textbf{Mentions} \\
\hline
\emph{Test\_large\_person}
    & standard &  0.84 &  0.8  &  0.82 &  1021 \\
    & fine-tuned &  0.77 &  0.99 &  0.87 &  1021 \\
\hline
\emph{Test\_small\_person}
    & standard  &  0.78 &  0.79 & 0.78 & 273 \\
    & fine-tuned & 0.69 &  0.95 &  0.8  &  273 \\
\hline
Internet news & standard & 0.84  & 0.94  & 0.89  & 324 
\end{tabular}
\label{tab:metrics_fine_tuned_ner}
\end{table}

\begin{table}[!ht]
\caption{Performance of the \emph{protagonistTagger} on various datasets.}
 \scriptsize
 \centering
 \setlength{\arrayrulewidth}{0.1mm}
\setlength{\tabcolsep}{3pt}
\renewcommand{\arraystretch}{1}
\begin{tabular}{ p{4cm} | p{1.5cm}| p{1.5cm}| p{1.5cm} }
 \textbf{Testing set} & \textbf{Precision} & \textbf{Recall} & \textbf{F-measure} \\
\hline
Novels - \emph{Test\_large\_names}    &    0.88 &     0.87 &        0.87 \\
Novels - \emph{Test\_small\_names}    &    0.83 &     0.83 &       0.83 \\
Internet news & 0.8  &     0.78 &        0.78 \\
\end{tabular}
\label{tab:full_tags_large_set_overall_metrics}
\end{table}

The \textit{protagonistTagger} tool achieves high results on all the tested novels -- precision and recall above 83\% (see~Table~\ref{tab:full_tags_large_set_overall_metrics}). The tool's performance on \emph{Test\_small\_names} shows that it can be successfully used for new distinct novels to create a larger corpus of annotated texts. The best proof of novels' diversity in the test sets is the tool's performance, whose precision varies from 79\% to even 96\% for different novels. Even though the performance is tested on various texts, the precision of the annotations remains high, proving the applicability of the proposed method in the literary domain. \par

The new dataset with internet news written in Polish contains around 1,000 sentences~\cite{pachocki2020categorization}. It is annotated with 100 identifiers (full names of popular Polish individuals, e.g. politicians, actors, researchers, etc.). The overall quality of the annotations is high (both recall and precision above 78\%). Even though, we applied only a few simple rules and no additional resources (e.g. dictionaries) in the NED phase.

\section{Conclusions}
In this paper, we propose a method and a tool for \textit{person} entity linkage in various domains -- \textit{protagonistTagger}. We also gathered datasets to express the problem of individuals' matching with text mentions. The method uses pretrained NER models and various techniques for NED. The initial tests were performed in the English literary domain. Most recent experiments prove the adaptability and effectiveness of the proposed approach in another language and domain of Internet news. The tool proved its effectiveness in achieving satisfactory results. The only precondition of using the tool is access to the predefined tags defining the full names (persons' identifiers) to be linked with named entities appearing in a text. A fascinating field of future applications is annotating texts from social media and using these annotations to investigate human opinions and analyzing sentiments. 

\bibliographystyle{splncs04}
\bibliography{custom}
\end{document}